\documentclass[preprint,12pt]{elsarticle}

\usepackage{caption}
\usepackage{subcaption}
\usepackage{amsmath,amssymb,amsfonts}
\usepackage[ruled,vlined]{algorithm2e}
\usepackage{graphicx}
\usepackage{textcomp}
\usepackage{xcolor}
\usepackage{multirow}
\usepackage[normalem]{ulem}
\usepackage{float}
\usepackage[utf8]{inputenc}
\useunder{\uline}{\ul}{}

\journal{Advanced Engineering Informatics}

\begin{document}

\begin{frontmatter}



\title{Multiagent Reinforcement Learning with Neighbor Action Estimation} 

\author[1]{Zhenglong Luo}

\author[1]{Zhiyong Chen\corref{cor1}}
\ead{zhiyong.chen@newcastle.edu.au}
\cortext[cor1]{Corresponding author.}

\author[2]{Aoxiang Liu}
\affiliation[1]{
  organization={School of Engineering, University of Newcastle},
  city={Callaghan},
  state={NSW},
  country={Australia}
}

\affiliation[2]{
  organization={School of Automation, Central South University},
  city={Changsha},
  state={Hunan},
  country={China}
}
\begin{abstract}

Multiagent reinforcement learning, as a prominent intelligent paradigm, enables collaborative decision-making within complex systems. However, existing approaches often rely on explicit action exchange between agents to evaluate action value functions, which is frequently impractical in real-world engineering environments due to communication constraints, latency, energy consumption, and reliability requirements.

From an artificial intelligence perspective, this paper proposes an enhanced multiagent reinforcement learning framework that employs action estimation neural networks to infer agent behaviors. By integrating a lightweight action estimation module, each agent infers neighboring agents' behaviors using only locally observable information, enabling collaborative policy learning without explicit action sharing. This approach is fully compatible with standard TD3 algorithms and scalable to larger multiagent systems.

At the engineering application level, this framework has been implemented and validated in dual-arm robotic manipulation tasks: two robotic arms collaboratively lift objects. Experimental results demonstrate that this approach significantly enhances the robustness and deployment feasibility of real-world robotic systems while reducing dependence on information infrastructure.

Overall, this research advances the development of decentralized multiagent artificial intelligence systems while enabling AI to operate effectively in dynamic, information-constrained real-world environments.

\end{abstract}


\begin{highlights}
\item Proposes a lightweight action estimation network for multiagent cooperation.
\item Enables agents to coordinate using only local observations without communication.
\item Enhances robustness under conditions of bandwidth constraints, latency, and high noise.
\item Reduces communication load and energy consumption requirements for deployed robotic systems.
\item Is validated on dual-arm manipulation with TD3, scalable to larger multi-agent teams.
\end{highlights}


\end{frontmatter}


\section{Introduction}

Reinforcement Learning (RL) has traditionally focused on single-agent settings, where the objective is to optimize the behavior of an individual agent. However, real-world applications often involve multiple agents that must interact, collaborate, or compete within complex and dynamic systems. This has driven significant advancements in Multiagent Reinforcement Learning (MARL), a field dedicated to developing intelligent agents capable of effective interaction, coordination, and adaptation in multiagent environments.

Early RL methods, such as Q-learning \cite{watkins1992q}, represent value functions using Q-tables, explicitly enumerating the Q-value for each state-action pair. Despite their simplicity, these tabular approaches suffer from scalability issues due to the combinatorial growth of states and actions, a problem further exacerbated in MARL scenarios, where multiple agents jointly contribute to an exponential expansion of the state–action space.

To address scalability, Deep Q-Networks (DQN) were introduced \cite{lecun2015deep}, replacing conventional Q-tables with neural networks to approximate Q-values. This advancement significantly improved generalization capabilities and computational efficiency, enabling RL algorithms to handle complex tasks with high-dimensional, continuous state–action spaces. DQN and its variants have achieved remarkable success across diverse domains, including industrialized construction, game playing, robotic control, and real-world decision-making problems \cite{Apolinarska2021,van2016deep,silver2017mastering,wang2018deep}.

Beyond value-based approaches, RL has expanded to include policy-based and hybrid algorithms. Policy Gradient (PG) methods directly optimize policies to maximize cumulative rewards, while Actor–Critic (AC) frameworks combine policy optimization with value function approximation, offering a balance between convergence speed and stability. A notable AC-based method, Twin Delayed Deep Deterministic Policy Gradient (TD3) \cite{fujimoto2018td3}, introduced key innovations, dual critic networks, delayed policy updates, and target action noise, that enhance policy stability and mitigate overestimation bias, achieving superior performance in complex continuous-control tasks. These algorithms have achieved considerable success in industrial construction as well, exemplified by Digital Twin–Driven Deep RL(DTDRL) \cite{Lee2022} for adaptive task allocation using PG-based Proximal Policy Optimization (PPO) to train task allocation policies, and Real-Time Twin DRL(RTTDRL) \cite{Tang2025} employing a DRL real-time scheduling optimization framework in a dual-cantilever crane system to enhance efficiency and reduce energy consumption.

Despite these advancements, adapting RL methods to MARL settings introduces additional challenges. One critical factor is the extent of information sharing among agents; for example, agents may operate under full, partial, or no information sharing schemes. Early MARL approaches often assumed full information exchange. Nash Q-learning \cite{hu2003nash} extended Q-learning to stochastic multiagent games by incorporating Nash equilibria into Q-value updates. While effective in discrete-state settings, it struggled to scale to large, continuous problems. This line of research was further advanced in \cite{luo2025multi}, which proposed a deep Q-network algorithm capable of learning multiple Q-vectors using Max, Nash, and Maximin strategies.  

Deep MARL methods subsequently emerged to address these complexities. Algorithms such as Multiagent Deep Deterministic Policy Gradient (MADDPG) \cite{lowe2017multi} and Counterfactual Multiagent Policy Gradient (COMA) \cite{foerster2018counterfactual} employ centralized critics for policy evaluation and decentralized actors for execution, substantially improving cooperative behavior and training stability. The Multi-agent PPO (MAPPO)algorithm, based on this framework, has been widely applied to optimize production line-logistics coordination scheduling in industrial assembly \cite{Li2025}. However, reliance on full communication incurs significant computational and informational overhead, limiting scalability.

To balance efficiency and optimality, partial information sharing methods have been proposed, enabling localized interactions among agents. Representative algorithms include Differentiable Inter-Agent Learning (DIAL) \cite{foerster2016dial}, Communication Network (CommNet) \cite{sukhbaatar2016commnet}, and graph neural network–based frameworks \cite{jiang2020graph}, which allow agents to aggregate and share neighborhood-level information. This strategy reduces communication costs and improves scalability, though it may come at the expense of global optimality due to restricted viewpoints.

In environments where communication is impractical or unavailable, no-information-sharing algorithms such as Independent Q-Learning (IQL) and Independent Policy Gradient (IPG) \cite{tan1993iql} enable agents to optimize their behaviors independently, treating other agents as part of the environment. However, independent training suffers from non-stationarity caused by constantly evolving agent policies, which undermines convergence and performance stability. Local observation–based methods encounter similar challenges, as reliance on limited observations increases the risk of local suboptimality.

To address non-stationarity and local optimality challenges in non-communicative scenarios, recent research has increasingly explored opponent modeling techniques. For instance, Learning with Opponent-Learning Awareness (LOLA) \cite{foerster2018lola} explicitly incorporates opponents’ policy gradients into an agent’s learning objective. By estimating and integrating anticipated opponent strategy updates, LOLA enables agents to proactively adjust their policies, thereby improving convergence in both competitive and cooperative settings.
   
Similarly, Deep Recurrent Q-Networks (DRQN) \cite{hausknecht2015deep} employ recurrent neural networks to capture temporal dependencies in partially observable settings. While effective at leveraging historical observation sequences to enhance policy performance, DRQN introduces greater model complexity and faces training challenges, particularly gradient instability during long-term sequence modeling.

However, in current industrial multiagent environments, a certain degree of information sharing remains necessary for multirobot collaboration. For instance, Model-Driven Contextual Reinforcement Learning (MCRL) \cite{Wang2025_MCRL} framework in Engineering Applications of Artificial Intelligence for collaborative robotic manipulation tasks was proposed. This approach enhances the generalization capability of underlying control strategies in uncertain environments by leveraging contextual information and probabilistic forward models, and its engineering effectiveness has been validated through multirobot simulations and dual-arm physical platforms. Nevertheless, this framework relies on system dynamics modeling and model prediction processes, which may limit its deployment flexibility and scalability in communication-constrained or decentralized scenarios.

Inspired by the strengths of opponent modeling approaches, particularly the proactive adaptation from LOLA, as well as the action estimation capabilities from recurrent models like DRQN, this paper proposes a novel estimation-based method, Action Estimation Network(AEN), specifically tailored for MARL scenarios without explicit inter-agent action communication.

We propose the AEN-TD3 algorithm, an extension of TD3 that incorporates an action estimation network to explicitly infer the actions of other agents. The estimation network predicts the actions of non-communicating agents using only locally available observations and historical information, thereby compensating for the absence of explicit communication. These predicted actions are combined with each agent’s own state–action pair and fed into the critic network, enhancing the accuracy of Q-value approximations in dynamic multiagent environments.

By explicitly modeling the actions of other agents, AEN-TD3 mitigates the fundamental MARL challenges of non-stationarity and partial observability, enabling more stable training and robust coordination without direct communication. Moreover, the approach preserves the computational efficiency and scalability of independent training methods, while avoiding the communication overhead inherent in full-information algorithms.

To rigorously validate our approach, we evaluate AEN-TD3 in a realistic and physically complex dual-arm manipulation task within the Mujoco simulation environment \cite{todorov2012mujoco}. Unlike conventional MARL benchmarks that often employ simplified, game-like scenarios, our experimental setting incorporates practical physical factors such as friction, gravity, and vibrations induced by joint interactions. Successful validation under these demanding conditions highlights AEN-TD3’s robustness and practical relevance to real-world MARL applications. 

\section{Methodology}

In this section, we introduce the main concept of the action estimation network, using the TD3 algorithm as the baseline architecture. The AEN is designed as a modular component that can be integrated with various RL algorithms. When combined with TD3, the resulting algorithm is referred to as AEN-TD3. This algorithm aims to facilitate effective learning in MARL scenarios where agents are unable to exchange action information with one another.

\subsection{Action Estimation Network}
\label{subsec:AEN}

A reinforcement learning algorithm updates the action-value function $Q(s, a)$ using the transition tuple $(s, a, r, s_{\text{next}}, a_{\text{next}})$, where $s$ is the current state, $a$ is the action taken under the current policy, $r$ is the immediate reward received, $s_{\text{next}}$ is the next state, and $a_{\text{next}}$ is the next action also selected using the current policy.  

In this study of a multiagent scenario, the reward $r$ is defined for a cooperative target. For example, in the `two arms lift a industrial component' task, the reward is determined by the height $r_{\rm height}$ and the posture-related term $r_{\rm angle}$ (defined as the negative tilt angle) of the object (e.g., a industrial component) being manipulated, i.e.,
$r = r_{\rm height} + r_{\rm angle}$. The robots cooperate to maximize the accumulated reward.

The RL policy is learned by each agent independently in a decentralized manner. For agent $i$, its state is denoted by $s^i$, and the states of the remaining agents are denoted by $s^{-i}$. That is, the overall state can be written in compound form as $s = (s^i, s^{-i})$. Similarly, the action is decomposed as $a = (a^i, a^{-i})$, where $a^i$ is the action of agent $i$ and $a^{-i}$ denotes the actions of all other agents.

Agent $i$ has access to the reward $r$ and the full state $s$, but only to its own action $a^i$. It does not have access to the actions of other agents in communication-limited scenarios, particularly when the policy functions of those agents are unknown.
As a result, from the perspective of agent $i$, the action input to the critic becomes incomplete, leading to significant errors in Q-value estimation. To address this issue, we introduce the AEN module to estimate the actions of other agents, $a^{-i}$, as follows
\begin{align}
a^o = e_{\psi}(s^{-i})  
\end{align}
Here, $e_{\psi}$ is a neural network parameterized by the weight vector $\psi$.
In analogy to the critic and actor networks in TD3, AEN also maintains a target network, denoted as $e_{\psi^{\prime}}(s^{-i})$, which is parameterized by a separate weight vector $\psi^{\prime}$.

AEN-TD3 employs the standard Double Q-learning algorithm, which uses two independently trained critic networks, denoted as  $Q_{\theta_1}(s, a^i, a^o)$ and $Q_{\theta_2}(s, a^i, a^o)$, to estimate Q-values. These networks are parameterized by the weight vectors $\theta_1$ and $\theta_2$, respectively, and have corresponding target networks parameterized by $\theta_1^{\prime}$ and $\theta_2^{\prime}$. This approach is designed to mitigate overestimation bias in value estimation.
Due to the inclusion of the AEN module, the key difference between these critic networks and their conventional counterparts lies in the use of the estimated action $a^o$, instead of the actual joint action component $a^{-i}$, for Q-value estimation.

In a conventional centralized setting, the actor network is denoted by $a = \pi_\phi(s)$, where it generates the complete action vector for all agents. In contrast, under the decentralized setting considered here, the actor network for agent $i$ generates only its own action, given by $a^i = \pi_\phi(s^i)$. 
The actions of the other agents, $a^{-i}$, are estimated as $a^o$ by the AEN module. The actor network is parameterized by the weight vector $\phi$, and a corresponding target actor network is maintained with parameters $\phi^{\prime}$.

\subsection{Update of Networks}

The temporal difference (TD) target used for updating the critic networks is defined as follows: 
\begin{align}
y = r + \gamma \min_{j=1,2} Q_{\theta'_j}(s_{\text{next}},   \pi_{\phi^{{\prime}}}(s^i_{\text{next}}) + \tilde\epsilon, e_{\psi^{{\prime}}}(s^{-i}_{\text{next}})), \nonumber \\
\tilde\epsilon \sim \text{clip}(\mathcal{N}(0,\tilde{\sigma}), -c, c),
\label{eq:td3_q_target}
\end{align}
where $\gamma$ is the discount factor, and $s_{\text{next}} = (s_{\text{next}}^i, s_{\text{next}}^{-i})$ represents the next state. The noise term $\tilde\epsilon$ is sampled from a clipped Gaussian distribution to ensure that the perturbed action remains close to the target action, which stabilizes learning. Here, $\tilde{\sigma}$ denotes the standard deviation of the noise, and $c$ defines the clipping range. It is worth noting that the resulting actions  $\pi_{\phi^{{\prime}}}(s^i_{\text{next}}) + \tilde\epsilon$ and $e_{\psi'}(s^{-i}_{\text{next}})$ are also clipped to lie within the valid action range $[a_{\text{min}}, a_{\text{max}}]$. 

In \eqref{eq:td3_q_target}, the target actor network $\pi_{\phi^{\prime}}$ and the target AEN $e_{\psi^{\prime}}$ are used to generate the next predicted actions for agent $i$ and its neighbors, respectively. The target Q-value is then computed using the target critic networks $Q_{\theta'_j}$, for $j=1,2$. The loss for each critic network $Q_{\theta_j}$ ($j=1,2$) is defined as: 
\begin{align}
L(\theta_j) = \mathbb{E}_{(s, a^i, a^o, r, s_{\text{next}})\sim\mathcal{B}}\left[(Q_{\theta_j}(s, a^i, a^o) - y)^2\right], \label{loss}
\end{align}
where $\mathcal{B}$ denotes the replay buffer from which experience tuples are sampled. The critic network parameters $\theta_1$ and $\theta_2$ are updated by minimizing the corresponding loss functions.

To update the actor network, AEN-TD3 adopts the same delayed policy update strategy as TD3, where the policy network $\pi_{\phi}$ is updated less frequently than the critic networks. Specifically, the actor and its corresponding target network are updated once every $d$ critic updates, with $d$ typically set to 2. During each update, the objective is to maximize the Q-value estimated by the first critic network $Q_{\theta_1}$. Accordingly, the actor parameters $\phi$ are updated by applying the following gradient:
\begin{align}
\nabla_{\phi}J(\phi) = \mathbb{E}_{(s, a^i, a^o, r, s_{\text{next}})\sim\mathcal{B}} \left[\nabla_{\phi}Q_{\theta_1}(s,\pi_{\phi}(s^i), a^o))\right]. \label{actorpdate}
\end{align}

Since the AEN's task of estimating other agents' actions is essentially a form of policy approximation, its neural network structure is similar to that of the actor network. As a result, the action estimation network is updated using the same method as the actor network. Specifically, the AEN parameters $\psi$ are updated by applying the following gradient:
\begin{align}
\nabla_{\psi}J(\psi) =   \mathbb{E}_{(s, a^i, a^o, r, s_{\text{next}})\sim\mathcal{B}} \left[\nabla_{\psi}Q_{\theta_1}(s, a^i, e_{\psi}(s^{-i}))\right].
\label{AENupdate}
\end{align}

After updating the network parameters $\theta_1$, $\theta_2$, $\phi$, and $\psi$, the corresponding target networks are updated using a soft update mechanism:
\begin{align}
      &\theta'_j \leftarrow \tau \theta_j + (1 - \tau) \theta'_j,\; j=1,2 \nonumber\\
      &\phi' \leftarrow \tau \phi + (1 - \tau) \phi' \nonumber\\
      &\psi' \leftarrow \tau \psi + (1 - \tau) \psi' 
\label{targetupdate}
\end{align}
where $\tau \in (0, 1)$ is the soft update rate that controls the speed of target network tracking.

\begin{algorithm}[t]
\caption{AEN-TD3 Algorithm for Agent $i$}
\label{alg:td3}

\KwSty{Parameter Setting}: Total episodes $M$, episode length $T$, minibatch size $N$, noise parameters $\sigma$, $\tilde{\sigma}$, clipping bound $c$, soft update rate $\tau$, delay parameter $d$\;

\KwSty{Initialize networks}: Initialize the critic networks  $\theta_1$ and $\theta_2$, the actor network $\phi$, and the AEN $\psi$\;

\KwSty{Initialize target networks}: $\theta^{\prime}_1 \leftarrow \theta_1$, 
$\theta^{\prime}_2 \leftarrow \theta_2$, $\phi^{\prime} \leftarrow \phi$, $\psi^{\prime} \leftarrow \psi$\;

\KwSty{Initialize replay buffer} $\mathcal{B}$\;

\For{episode $= 1$ to $M$}{
  Observe initial state $s_1 =(s_1^i, s_1^{-i})$ \;
  \For{$t = 1$ to $T$}{
    Select action with exploration noise: $a^i_t = \pi_{\phi}(s^i_t) + \epsilon$, where $\epsilon \sim \mathcal{N}(0,\sigma)$\;
    Compute estimated action:
    $a_t^o = e_{\psi}(s_t^{-i})$\;
    Execute action $a^i_t$, observe reward $r_t$ and next state 
    $s_{t+1} =(s_{t+1}^i, s_{t+1}^{-i})$\;
    Store transition $(s_t, a^i_t, a^o_t, r_t, s_{t+1})$ in $\mathcal{B}$\;
    Sample a minibatch of $N$ transitions $(s, a^i, a^o, r, s_{\text{next}})$ from $\mathcal{B}$\;
    Calculate the TD target $y$ for each sample according to
    \eqref{eq:td3_q_target}\;

    Compute the empirical mean of the loss \eqref{loss}:
    $L(\theta_j) = \frac{1}{N} \sum (y - Q_{\theta_j}(s, a^i, a^o))^2$, $j=1,2$\;

    Update the critic networks $\theta_1$ and $\theta_2$ by minimizing their corresponding losses\;

    \If{$t \mod d = 0$}{ 
    
    Update the actor network $\phi$ by applying the empirical gradient of \eqref{actorpdate}:
      $\nabla_{\phi} J(\phi) = \frac{1}{N} \sum\nabla_{\phi}Q_{\theta_1}(s,\pi_{\phi}(s^i), a^o))$\;

      Update the AEN $\psi$ by applying the empirical gradient of \eqref{AENupdate}:
      $\nabla_{\psi} J(\psi) = \frac{1}{N} \sum \nabla_{\psi}Q_{\theta_1}(s, a^i, e_{\psi}(s^{-i}))$\;

      Update the target networks according to \eqref{targetupdate}.
    }
  }
}
\end{algorithm}

With the network architecture and update rules introduced above, the complete algorithm,
AEN-TD3, is summarized in Algorithm~\ref{alg:td3}, which incorporates the AEN module into the standard TD3 framework. This integration enables effective estimation of other agents' policies in the absence of inter-agent communication. As a result, the algorithm improves the accuracy of Q-value estimation, despite the lack of explicit information exchange. Consequently, AEN-TD3 achieves performance comparable to centralized methods while greatly reducing the reliance on communication.

\section{Experimental Results}
\label{sec.Experiments}

To evaluate the effectiveness of AEN-TD3 on a cooperative task, we compared its performance with that of conventional TD3 implemented in a centralized manner, which serves as the baseline. Experiments were conducted using the Robosuite environment \cite{zhu2020robosuite} on the task `two arms lift a industrial component'. Robosuite is a modular simulation framework built on the MuJoCo physics engine \cite{todorov2012mujoco}, offering a range of pre-built simulation environments for robotic manipulation. After training, both the TD3 and AEN-TD3 policies were deployed on a physical robotic platform to evaluate their real-world performance. The two robotic arms used in the task are UR5e models.

\subsection{Simulation Environment}

In the `two arms lift a industrial component’ Robosuite environment, two robotic arms are used, forming a multiagent setting. For clarity, the first arm ($i = 1$) is referred to as the left arm, and the second arm ($i = 2$) as the right arm. The primary objective is for both arms to lift a centrally positioned industrial component simultaneously by coordinating their joint movements while keeping the industrial component as stable as possible during the process. The state $s^i\in {\mathbb R}^{6}$ represents the joint angles of the $i$-th robotic arm. 
The reward $r$ is defined in Section~\ref{subsec:AEN}.

We adopt {\it `Joint Velocity'} as the control action. Specifically, actions are applied to two joints of each robotic arm, resulting in an action space of $a^i \in {\mathbb R}^3$ per arm. The parameter that determines the frequency of the input signal is set to {\it `control frequency=4'}.  A high control frequency increases the number of steps needed to reach a target, which complicates the learning process and can lead to unstable policy training. To address this, we use a lower control frequency 4~Hz. However, this creates a mismatch with the real robotic system, which operates at 20~Hz. Applying a low-frequency policy to the real system may cause abrupt accelerations, resulting in oscillations in the robotic arms. This issue will be addressed in a later section.

The parameter that defines the end of the current cycle, after a specified number of actions have been executed and the environment reinitialized, is set to {\it `horizon=200'}. This setting allows the two robotic arms to raise the industrial component to its highest position while following the ideal path, thereby supporting effective training completion.

We also define an early termination condition that may end an episode before the full horizon is reached, based on the detection of unsafe mechanical configurations. This mechanism is introduced to ensure safe policy transfer to real robotic systems. Let $d_t$ denote the distance between the two grippers at time $t$, and $d_0$ the initial distance when the grippers symmetrically clamp the component handles. A safe deviation threshold $\delta$ is defined, and early termination is triggered if $|d_t - d_0| > \delta$.

This early termination condition effectively prevents the agent from learning harmful force-exerting strategies, such as excessive pulling or squeezing, which could cause structural failure during real-world deployment, even if such actions merely lead to the industrial component slipping in simulation. Therefore, the early termination condition encourages the learning of safe and reliable policies suitable for deployment in physical environments.

\subsection{Simulation Results and Evaluation}

The comparison between AEN-TD3 in a decentralized setting and conventional TD3 implemented in a centralized manner is based on simulation results, as discussed in this section. The comparable performance verifies the effectiveness of AEN-TD3.

\subsubsection{Centralized Setting: TD3}

Both robotic arms have full access to each other’s information, including state, action, reward, and replay buffer, effectively functioning as a single composite system. The control policy is trained in a centralized manner using this shared information. The environment is modeled with a complete state space $s\in {\mathbb R}^{12}$, representing the joint angles of both robotic arms, and a complete action space $a \in {\mathbb R}^6$, representing the control signals for all six joints. This formulation enables the use of conventional single-agent RL algorithms.

Specifically, we use the TD3 algorithm to train the model. For the critic network, the input is the concatenation of the 12-dimensional overall state vector and the 6-dimensional overall action vector, forming an 18-dimensional input. This is passed through two fully connected layers with ReLU activation functions, and the final output is a scalar representing the approximate Q-value, which is used for network updates.

The actor network takes as input a 12-dimensional overall state vector, which is processed through two fully connected layers of 256 units each, with ReLU activation functions in between. The output then passes through a Tanh activation function, combined with additional operations, to constrain the values to the range $[-0.04, 0.04]$. The resulting 6-dimensional overall action vector is directly fed into the simulation environment as the action values.

\begin{figure}
    \centering
    \includegraphics[width=0.8\columnwidth]{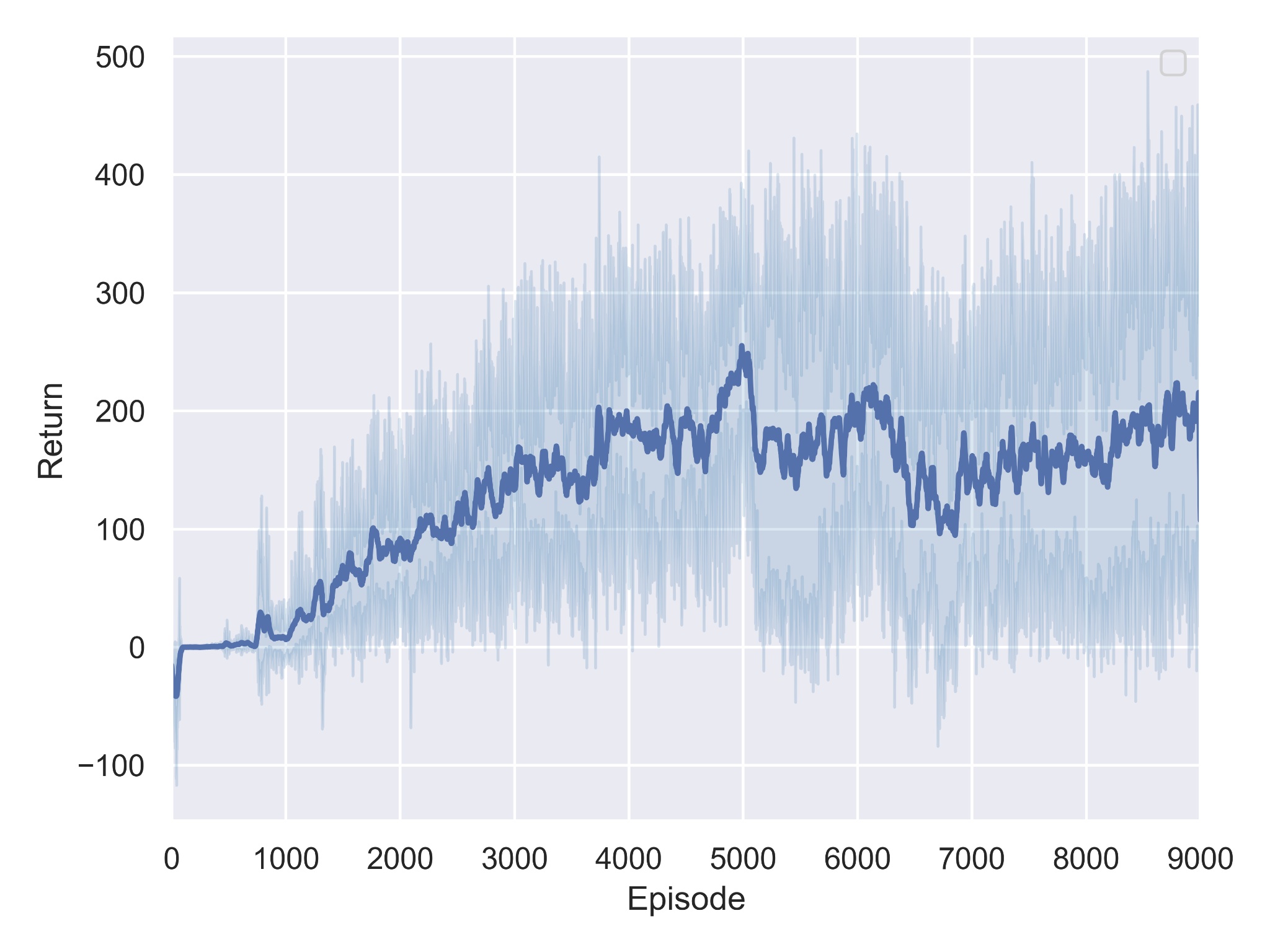}
\caption{Performance of returns during TD3 training.}
    \label{TD3SimReturngood}
\end{figure}

During training with the original TD3 algorithm, we conducted ten independent experiments. Under identical environmental conditions, the parameters of the actor and critic networks were randomly initialized at the start of each 9,000-episode training session. Of these ten runs, eight successfully achieved the task of smoothly raising the component to a height exceeding $1.3$-$1.4$ m without causing deformation or compression, as illustrated in Figure~\ref{TD3SimReturngood}. These trials were therefore classified as successful.

\begin{figure}
    \centering
    \includegraphics[width=0.8\columnwidth]{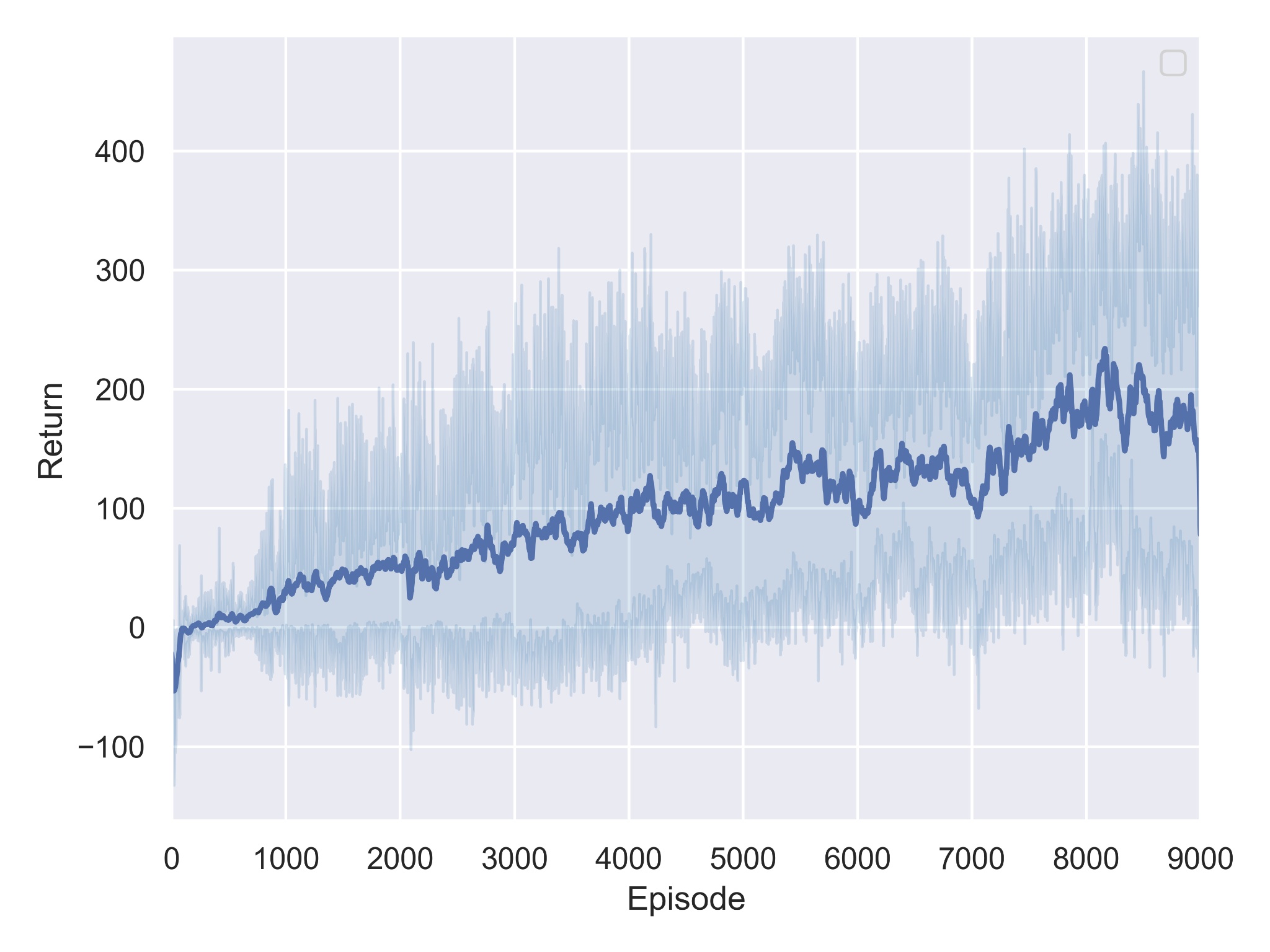}
\caption{Performance of returns during AEN-TD3 training.}
\label{ESTSimReturngood}
\end{figure}

A noticeable decline is observed in the figure, primarily caused by a drop in one particular training trial at this stage. Nevertheless, this run eventually self-corrected and successfully resumed convergence toward the correct policy.
Two experiments, however, failed to reach comparable return levels and were deemed unsuccessful. We attribute this failure mainly to two factors: (i) insufficient exploration of optimal trajectories during the random exploration phase, and (ii) the accumulation of poor-quality data during random sampling, where certain actions initially appeared reasonable but ultimately led to the component being dropped.

Finally, we evaluated the trained policy in an environment with both initialization and exploration noise disabled. Under its control, the dual-arm system successfully lifted the component to a height of 1.4 m, closely matching the target position, and completed the task within a single episode using 200 actions. These results demonstrate that the original TD3 algorithm can effectively accomplish the specified task.

\subsubsection{Decentralized Setting: AEN-TD3}
 
In this decentralized setting, the two robotic arms do not exchange actions. Each arm updates its actions and optimizes its strategy solely based on locally available information, which includes the opposite arm’s position data (joint angles) and component-related information (height and tilt angle). Using this information, each arm independently trains with the AEN-TD3 algorithm. 

For each AEN-TD3 instance, the internal structure of the critic network remains unchanged; however, its input is formed by concatenating multiple vectors: the arm’s own state, the opposite arm’s state, the arm’s own action, and the opposite arm’s action estimated by the AEN, resulting in a total of 18 input dimensions.

The actor network retains its original activation functions, and the input is the arm’s own 6-dimensional state vector. Since training is independent for each arm, the size of every network layer is reduced by half compared to the previous case, resulting in 128 units per layer.

The AEN has the same structure as the actor network: two fully connected layers with ReLU activation in between, followed by a Tanh activation and additional scaling functions to constrain the output to
$[-0.04, 0.04]$. Its input is a 6-dimensional state vector of the opposite arm, sampled from the replay buffer, and its output is a 3-dimensional estimate of the opposite arm’s action.

Similar to the original TD3 algorithm, AEN-TD3 was trained over ten independent experiments. Compared to TD3, the training process of AEN-TD3 was slower, particularly in the early stages, where predefined safety constraints were frequently violated, leading to early termination. This can be attributed to the increased difficulty of coordinated exploration without access to the accurate actions of the other arm.

As in the centralized setting, eight out of the ten training runs successfully learned effective policies within 9,000 episodes, as shown in Figure~\ref{ESTSimReturngood}. The remaining two runs were recorded as failures, either due to prolonged convergence times of up to 14,000 episodes or no convergence.

Evaluation of the learned policies demonstrated successful completion of the lifting task, raising the component to a height exceeding 1.4 meters, comparable to the results obtained in the centralized setting. These comparable results validate the effectiveness of the proposed AEN-TD3 algorithm in a decentralized setting.

\subsection{Policy Deployment and Experiments}

While policies trained in simulation environments such as Robosuite have shown promising performance, directly transferring these strategies to physical robotic hardware presents substantial challenges. Discrepancies between simulated and real environments, such as differences in dynamics, actuation frequencies, and mechanical constraints, can significantly impair the real-world applicability of trained policies.

To address these challenges, we conducted a series of experiments in which our TD3-based policies, trained in simulation, were deployed on a physical robotic platform. This subsection describes the policy transfer methodology, the experimental setup on the real robotic arms, and a comparative analysis of the results, thereby demonstrating the feasibility and practical effectiveness of the proposed approach.

The physical robotic platform utilizes the same UR5e robot model as in the simulation. Furthermore, both the mounting positions and the initial joint configurations of the robotic arms are kept identical to those used in the simulated environment.

\subsubsection{Signal Interpolation}

In simulation, the policy was trained with a control frequency of $4$~Hz to promote stable learning. In contrast, the physical robotic system requires a control frequency of $20$~Hz to achieve smooth and responsive actuation. This mismatch in control frequencies can cause temporal inconsistencies and degraded performance when transferring policies from simulation to the real world.

\begin{figure}[H]
\centering
\includegraphics[width=0.7\columnwidth]{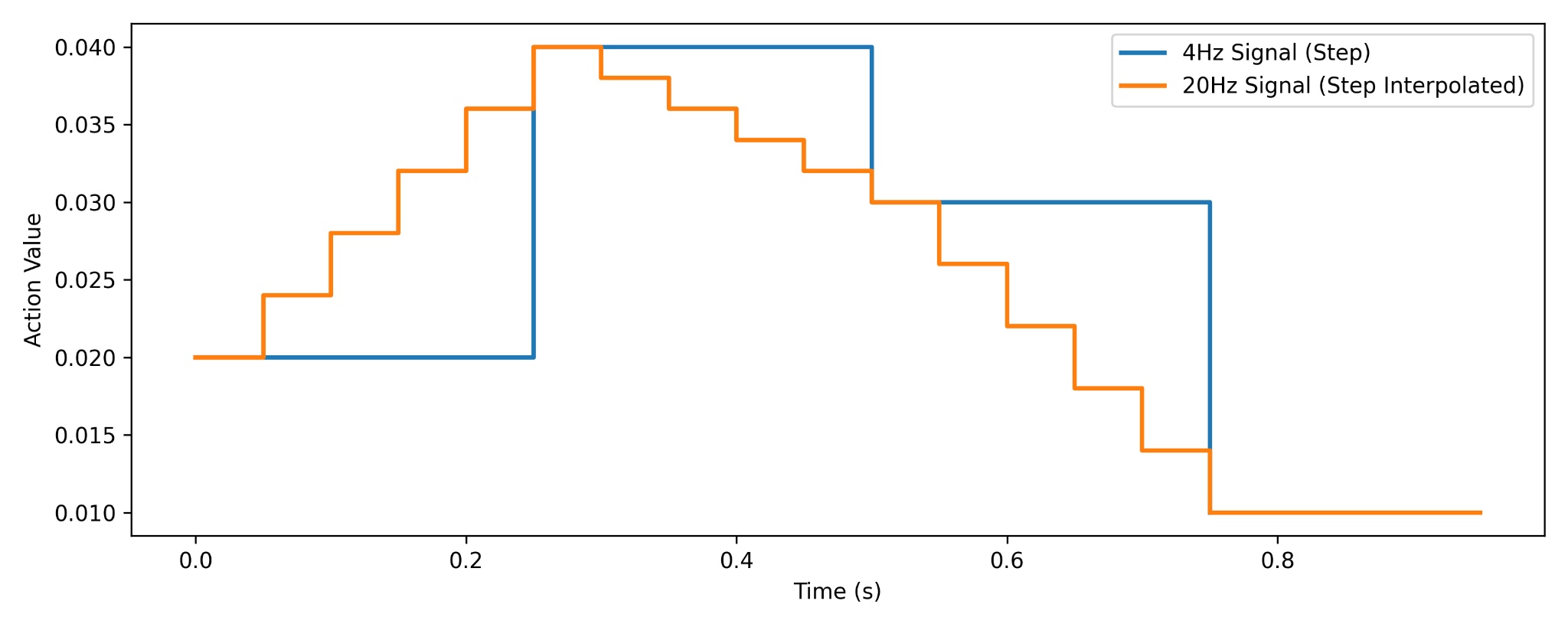}
\caption{Signal Interpolation Mechanism}
\label{SCM} 
\end{figure}

To overcome this discrepancy, we introduce a signal interpolation mechanism that maps low-frequency action outputs to high-frequency control commands, as illustrated in Figure~\ref{SCM}. Specifically, each action generated at $4$~Hz is extended over five consecutive $20$~Hz control cycles. Within this extension, linear interpolation is applied to ensure temporal continuity between successive actions. This approach satisfies the real-time constraints of the physical system, mitigates abrupt changes in control signals, and reduces oscillatory behavior, thereby improving the robustness of policy transfer from simulation to hardware.

Furthermore, even with smoothing operations, certain transitions between control signals may still exhibit large differences. Such abrupt changes can cause unsafe oscillations in the robotic arm due to inertial effects. To address this, an acceleration-based safety mechanism is implemented during real-world deployment. Specifically, if the change in the control signal between consecutive steps exceeds a predefined safety threshold, the signal is classified as unsafe and excluded from execution. This safeguard ensures stable and safe operation of the robotic system.

\subsubsection{Tightened Safety Constraints}

A safety margin is imposed to constrain the movement range of the robotic grippers, thereby limiting the force exerted on the component handles and preventing potential tearing or crushing. During the simulation phase, this threshold was set to $\delta = 0.02$ corresponding to an allowable horizontal displacement of $2$~cm. Rendering results indicate that this constraint effectively restricted the end-effector’s three-dimensional linear force to below $30$~N. 

However, during real-world deployment, we identified a structural discrepancy between the simulation and the physical hardware. In particular, Robosuite omits the coupler component in its UR5e gripper model, resulting in a mismatch between the simulated and actual configurations. Consequently, policies that perform well in simulation may still cause excessive squeezing when applied to the physical system.

Given the relatively small size of the coupler, retraining the policy from scratch would be inefficient. Instead, we address the issue by tightening the safety constraint: specifically, the parameter  $\delta$ is reduced to further limit the gripper’s operational range, thereby compensating for the modeling discrepancy. Using the pre-trained policy as a baseline, we conduct an additional 200,000 training steps under the adjusted safety settings, with $\delta = 0.015$ and $\delta = 0.01$,  respectively. Experimental results show that this adjustment effectively prevents mechanical damage during real-world execution and highlights the robustness of our simulation-to-reality transfer approach.

\subsubsection{Results and Evaluation}

\begin{figure}
    \centering
    \includegraphics[width=0.7\columnwidth]{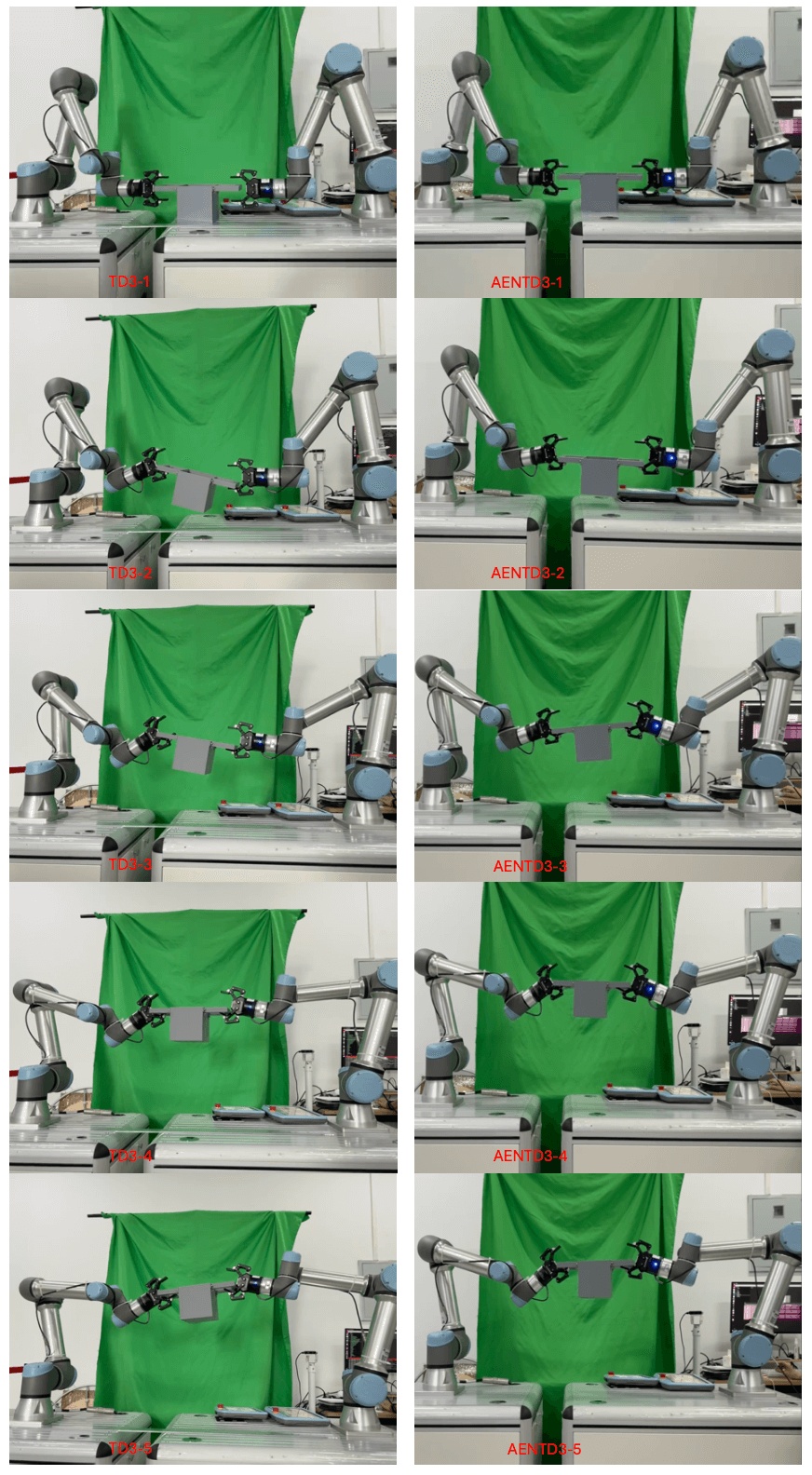}
\caption{Sequential snapshots of experiments on the physical robotic platform using policies trained with TD3 (left) and AEN-TD3 (right).}
    \label{Snapshots}
\end{figure}

\begin{figure}
    \centering
    \includegraphics[width=0.6\textwidth]{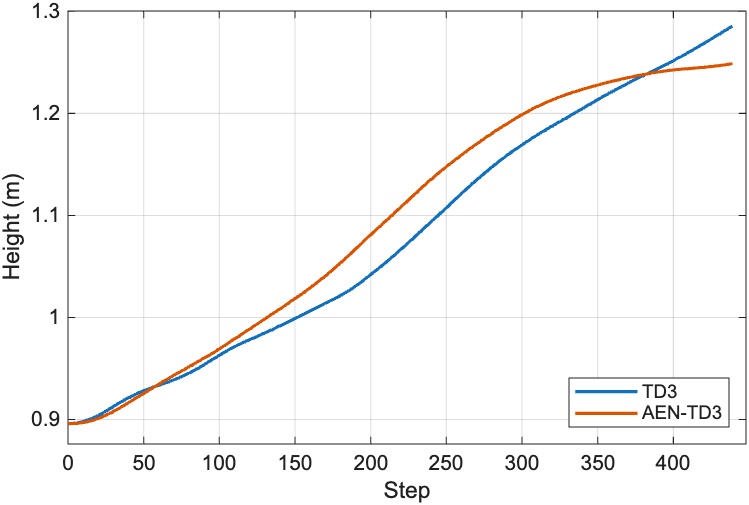}
\caption{Component height trajectories using policies trained with TD3 and AEN-TD3.}
    \label{TD3Real}
\end{figure}

When the final trained TD3 and AEN-TD3 policies were deployed on the physical robotic platform, the lifting process in the real environment lasted approximately 24 seconds. To illustrate the progression, snapshots were taken every 5 seconds, as shown in Figure~\ref{Snapshots}. The left and right columns correspond to the execution results of the TD3 and AEN-TD3 policies, respectively. Both strategies successfully lifted the component while maintaining stability throughout the process, with no instances of squeezing or tearing. Figure~\ref{TD3Real} further depicts the component height trajectories achieved by both policies in the real environment, providing a quantitative basis for performance comparison.

These experimental results demonstrate that, even when the two robotic arms operate as independent agents without access to each other’s actions, the AEN can effectively compensate for the missing information. Consequently, the AEN-TD3 algorithm achieves performance comparable to TD3 with full information sharing. This validates both the effectiveness and feasibility of the AEN-TD3 algorithm for real-world deployment and further confirms its successful performance.

\section{Conclusion}
\label{sec.Conclusion}

This paper proposes a TD3-based cooperative learning method enhanced with an action estimation network to address multiagent environments without direct action exchange. The approach enables two robotic arms to collaboratively lift objects while achieving performance comparable to centralized TD3 algorithms. Its effectiveness and practical feasibility are further validated through experiments conducted on physical robot platforms under real-world operational conditions. By reducing reliance on explicit communication, this framework provides a practical solution for collaborative robotic systems in information-constrained environments. Future work will focus on: extending the action estimation method to larger-scale multiagent systems, integrating complementary game strategies, and adapting the framework for non-collaborative and mixed-interaction tasks.

\section*{Funding sources}

This research did not receive any specific grant from funding agencies in the public, commercial, or not-for-profit sectors.

\bibliographystyle{elsarticle-harv}
\bibliography{Reference}

\end{document}